\documentclass[sigconf,nonacm]{acmart}

\usepackage{graphicx}
\usepackage{booktabs}
\usepackage{multirow}
\usepackage{amsmath}
\usepackage[table]{xcolor}
\usepackage{array}
\usepackage{tabularx}
\usepackage{subcaption}
\usepackage{enumitem}
\usepackage[most,breakable]{tcolorbox}

\definecolor{phcolor}{RGB}{180,0,0}
\definecolor{lightgray}{RGB}{240,240,240}
\definecolor{darkblue}{RGB}{31,73,125}
\definecolor{medblue}{RGB}{68,114,196}
\definecolor{lightblue}{RGB}{189,215,238}
\definecolor{darkgreen}{RGB}{55,96,35}
\definecolor{accentorange}{RGB}{197,90,17}

\newcommand{\sysname}{ARCagent}

\title{\sysname: An Adaptive Retrieval Calibration Agent for Clinical Question Answering}

\author{Yuyan Chen}
\affiliation{\institution{}\country{}}

\begin{abstract}
In diseases where clinical guidelines are incomplete, contested, or mutually contradictory, knowledge completeness and dynamic conflict-aware synthesis are two safety-critical properties that standard Retrieval-Augmented Generation systems do not provide.
Therefore, we present \sysname, an adaptive retrieval calibration clinical question-answering agent for ME/CFS, a disease where diagnostic frameworks coexist and major guidelines actively contradict each other on treatment.
\sysname\ contributes three components.
First, a 1,706-chunk, 10-source knowledge base with a structured inter-guideline conflict registry spanning all active ME/CFS diagnostic frameworks.
Second, a conflict-aware retrieval calibration pipeline that re-ranks retrieved evidence using query-specific focus and conflict signals.
Third, a benchmark scored by LLM-as-Judge, avoiding systematic underestimation averaging 10.1 percentage points caused by keyword matching.
\sysname\ achieves 95.3\%, outperforming all base LLMs.
Code is available at \url{https://github.com/Yukyin/ARCagent}.
\end{abstract}

\keywords{Retrieval-Augmented Generation, Clinical Question Answering, Conflicting Guidelines, ME/CFS}

\begin{document}

\maketitle

\section{Introduction}
\label{sec:intro}

Retrieval-augmented generation systems fail silently in domains where authoritative sources actively contradict each other: returning the most relevant chunk without flagging inter-source disagreement risks surfacing clinically harmful outputs \citep{gao2023ragsurv}. We identify \textit{inter-guideline conflict} as a first-class retrieval problem and propose \textit{conflict-aware retrieval calibration} as a general paradigm for handling it. The paradigm requires two properties. The first is \textit{knowledge completeness}: all coexisting and superseded guideline versions must be represented so that no stance is structurally suppressed. The second is \textit{dynamic conflict-aware synthesis}: retrieved evidence must be re-ranked and surfaced according to a structured conflict registry that encodes which sources disagree and on what claims, rather than leaving conflict detection to the language model at generation time.

ME/CFS instantiates this class of domains with unusual clarity. Six active diagnostic frameworks published between 1991 and 2021 coexist, and their treatment guidelines actively contradict each other: NICE NG206 \citep{nice2021ng206} prohibits graded exercise therapy and CBT as primary treatments, reversing NICE CG53 \citep{nice2007cg53}, while IQWiG N21-01 \citep{iqwig2023n2101} reports short-term CBT benefit based on a divergent appraisal of the same PACE trial data \citep{white2011comparison}. Any system that retrieves from either source without flagging this conflict risks clinically harmful outputs, making ME/CFS a tractable and high-stakes testbed for the paradigm.

Existing approaches address complementary but insufficient subsets of this problem. Standard Retrieval-Augmented Generation (RAG) \citep{lewis2020rag} conditions generation on retrieved passages and reduces hallucination \citep{guu2020realm}, but returns the most relevant chunk without detecting inter-source contradiction \citep{gao2023ragsurv}. Domain-adapted medical LLMs \citep{singhal2023medpalm,yang2023huatuogpt,nori2023capabilities} encode clinical knowledge but assume a static, non-contradictory knowledge base and cannot handle guideline version conflicts. Active retrieval methods such as Self-RAG \citep{asai2024selfrag} reduce compounding errors through sentence-level retrieval decisions \citep{shi2023replug}, yet provide no structured source-conflict annotation. None of these approaches treats conflict detection as a retrieval objective.

We present \sysname, which instantiates conflict-aware retrieval calibration on ME/CFS. \sysname\ makes three contributions. First, we propose a conflict registry architecture that encodes inter-guideline disagreements as structured annotations at the chunk level, enabling deterministic conflict detection and dual-attribution generation without relying on the language model to infer contradictions. Second, we construct MECFS-KB, a 1,706-chunk, 10-source knowledge base with a four-zone conflict registry, and release a 1,200-query evaluation benchmark across eight clinically motivated categories covering conflict surfacing, scope boundary enforcement, and revision history reasoning. Third, we show that keyword-based evaluation underestimates all models by an average of 10.1 percentage points in this domain, establishing LLM-as-Judge as the appropriate evaluation protocol when clinical paraphrasing is expected. Evaluated on the benchmark, \sysname\ achieves 95.3\%, outperforming all nine baseline models, with a 6.0 percentage point gain over its retrieval-first variant and a 17.5 percentage point gain over the best domain-specific medical LLM.

\section{Related Work}
\label{sec:related}

\paragraph{Retrieval-Augmented Generation.}
\citet{lewis2020rag} establish the RAG paradigm, with active retrieval methods such as FLARE \citep{jiang2023flare}, IRCoT \citep{trivedi2023ircot}, and Self-RAG \citep{asai2024selfrag} interleaving retrieval with generation to reduce multi-step errors, while \citet{gao2023ragsurv} identify knowledge currency and source conflict as open challenges. \citet{izacard2021leveraging} show that fusing evidence across multiple retrieved passages with a sequence-to-sequence reader substantially improves open-domain QA, and \citet{ram2023incontext} demonstrate that prepending retrieved passages as in-context examples without fine-tuning yields strong retrieval-augmented performance. \citet{shi2023replug} treat the language model as a black box and train a retrieval model to improve its perplexity, showing that retrieval can be tuned independently of the generator. \citet{guu2020realm} propose pretraining with a latent document index, demonstrating that retrieval-aware pretraining outperforms augmenting a frozen model at inference time.
\sysname\ addresses both by treating conflict detection as a first-class retrieval objective coupled with a structured conflict registry.

\paragraph{Clinical NLP and Evaluation.}
\citet{singhal2023medpalm} show that large LLMs encode substantial clinical knowledge but evaluate against a static, non-contradictory knowledge base, and domain-adapted models \citep{yang2023huatuogpt,lehman2023clinical} improve factual accuracy without handling guideline version conflicts. \citet{nori2023capabilities} evaluate GPT-4 on medical licensing examinations and show that general-purpose LLMs approach specialist performance without clinical fine-tuning, while \citet{jin2019pubmedqa} introduce a biomedical QA benchmark that requires reasoning over PubMed abstracts to answer yes/no/maybe questions, highlighting the gap between retrieval and reasoning in clinical NLP. \citet{agrawal2022large} study few-shot clinical NLP and find that prompt engineering closes much of the gap with supervised fine-tuning on EHR tasks, and \citet{pampari2018emrqa} construct a large-scale clinical QA dataset from EHR data showing that clinical QA requires domain-specific evidence retrieval beyond general open-domain corpora.
Building on \citet{zheng2023judging}'s LLM-as-Judge framework, we show that keyword matching underestimates all models in this domain, validating semantic evaluation when clinical paraphrasing is expected.

\section{Knowledge Base Construction}
\label{sec:kb}
\begin{figure*}[t]
  \centering
  \includegraphics[width=0.92\linewidth]{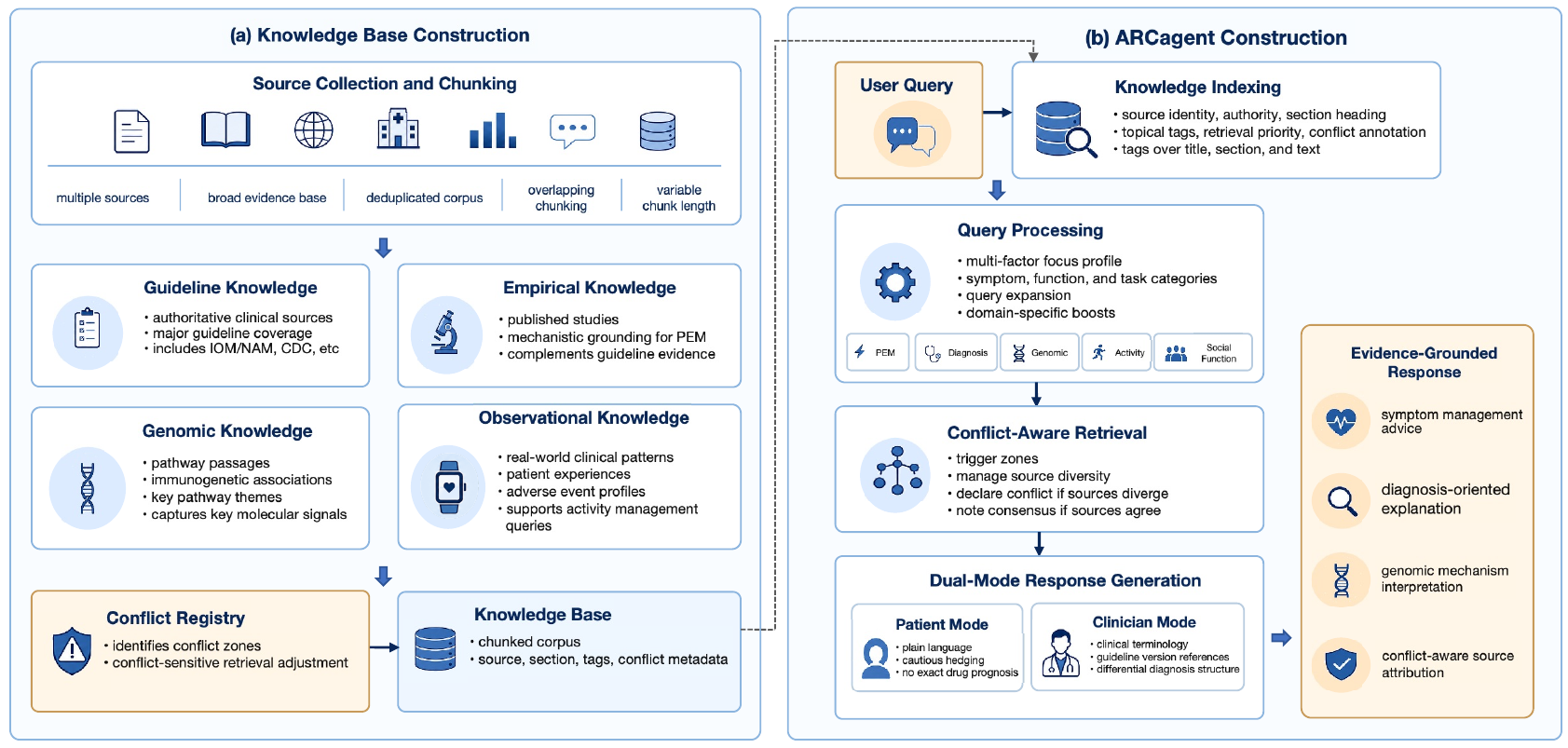}
    \caption{Overview of knowledge base and \sysname\ construction. (a) Knowledge base construction: sources are collected and chunked into guideline, empirical, genomic, and observational knowledge, with a conflict registry annotating conflict zones. (b) \sysname\ construction: a query is processed into a focus profile, resolved through conflict-aware retrieval, and answered via dual-mode response generation.}
  \label{fig:overview}
\end{figure*}

The knowledge base comprises 1,706 chunks totalling 263,587 words from 10 sources after MD5-hash deduplication, constructed via the pipeline shown in Figure~\ref{fig:overview}(a).
Chunking targets 350 words per chunk with 50-word overlap \citep{chen2017reading}, a minimum of 80 words, and a maximum of 600 words.
Table~\ref{tab:kb} summarises source composition.

\subsection{Guideline Knowledge}
Guideline knowledge accounts for 1,688 chunks from eight authoritative clinical sources, with IOM/NAM 2015 \citep{iom2015mecfs, stussman2020characterizing} as the largest contributor at 929 chunks, followed by CDC ME/CFS \citep{cdc2021mecfs}, NICE NG206 \citep{nice2021ng206}, the ME/CFS Clinician Coalition toolkit \citep{mecfscc2021}, IACFS/ME primer \citep{friedberg2014primer}, IQWiG N21-01 \citep{iqwig2023n2101}, CCC 2003 \citep{carruthers2003ccc}, and ICC 2012 \citep{carruthers2011icc}.
Empirical knowledge accounts for the remaining 18 chunks; although a small fraction by count, ablation results in Section~\ref{sec:ablation} confirm it provides mechanistic grounding for PEM-related queries that no guideline source substitutes.
An inter-guideline conflict registry encodes four substantive conflicts between coexisting guidelines.
The highest-severity entry captures the IQWiG N21-01 and NICE NG206 disagreement on CBT \citep{white2011comparison}.
Conflict passages receive a suppression penalty of $-2.0$ during standard queries and a surfacing reward of $+1.0$ when users explicitly request conflict information.

\begin{table}[t]
\centering
\caption{MECFS-KB source composition. IQWiG N21-01 chunks are annotated as conflicting with NICE NG206.}
\label{tab:kb}
\resizebox{0.95\linewidth}{!}{%
\begin{tabular}{llrr}
\toprule
\textbf{Source} & \textbf{Organisation} & \textbf{Chunks} & \textbf{Words} \\
\midrule
IOM/NAM 2015        & US NAM   & 929 & 109,145 \\
CDC ME/CFS          & US CDC                           & 215 &  20,529 \\
NICE NG206          & UK NICE                          & 214 &  24,285 \\
Clinician Coalition & ME/CFS CC, USA                   & 102 &  35,700 \\
IACFS/ME Primer     & Friedberg et al.                 &  82 &  24,464 \\
IQWiG N21-01 & IQWiG, Germany               &  58 &  17,219 \\
CCC 2003            & Carruthers et al.                &  51 &  15,056 \\
ICC 2012            & \textit{J Intern Med}            &  37 &  10,889 \\
Genomic             & GSE16059 &  17 &   5,950 \\
Clinical            & Rekeland et al.\ 2022            &   1 &     350 \\
\midrule
\textbf{Total}      &                                  & \textbf{1,706} & \textbf{263,587} \\
\bottomrule
\end{tabular}
}
\end{table}

\subsection{Genomic Knowledge Construction}
Seventeen genomic passages are constructed from a re-analysis of GSE16059, a gene expression dataset of 32 monozygotic twin pairs discordant for ME/CFS \citep{subramanian2005gsea}; the monozygotic design eliminates genetic confounding \citep{bell2013twins}, making expression differences attributable to disease state.
Per-probe paired Wilcoxon signed-rank tests on 54,675 probes with BH-FDR correction \citep{benjamini1995controlling} were mapped to 20,788 unique genes and ranked for GSEA \citep{subramanian2005gsea}, confirmed by Fisher's exact over-representation analysis.
Four passages encode the key pathway findings, ISG upregulation (enrichment score 0.841), OXPHOS downregulation ($-0.729$), HPA axis downregulation ($-0.476$), and the monozygotic causal inference rationale, with pathways above an absolute enrichment score of 0.30 receiving a $1.5\times$ retrieval boost; the ISG and OXPHOS findings are consistent with broader immune dysregulation evidence in ME/CFS \citep{montoya2017cytokine}.

\subsection{Clinical Observational Knowledge}
One clinical passage is derived from a re-analysis of wearable sensor data from \citet{rekeland2022wearable} across 27 ME/CFS patients over 168 days \citep{stussman2020characterizing}, defining crash events as step-count drops below 50\% of the seven-day rolling median for at least two consecutive days.
The re-analysis identified 68 crash events averaging 2.5 per patient, with depth 70--82\% below baseline and median duration of two days; a floor effect in the severe subgroup makes fixed-threshold detection unreliable, triggering retrieval for activity management queries where guideline text provides no equivalent characterisation.

\section{\sysname\ Construction}
\label{sec:arc_sys}

\sysname\ combines a structured knowledge indexing layer with a conflict-aware retrieval pipeline and a dual-mode generation stage, as illustrated in Figure~\ref{fig:overview}(b).

\subsection{Knowledge Indexing}
Each of the 1,706 chunks in MECFS-KB is stored with metadata encoding its source identity, authority, section heading, topical tags, retrieval priority, and conflict annotation, and is indexed with BM25Okapi \citep{robertson2009bm25, chen2017reading} over a merged field of title, section, text, and tags.
Tags are assigned in two passes, an initial ingestion-time pass using source-level keywords and a retroactive pass that scans chunk text for GET, CBT, PEM, orthostatic, and conflict-relevant terminology to correct under-tagged chunks, while conflict annotation is set explicitly for IQWiG N21-01 and NICE NG206 chunks on GET and CBT \citep{johnson2016mimic} and for all six diagnostic framework sources on diagnostic criteria, enabling the conflict registry to operate at the chunk level rather than inferring conflicts at query time.

\subsection{Query Processing and Conflict-Aware Retrieval}
On each query, \sysname\ classifies the query into a 13-dimensional binary focus profile spanning symptom domains (PEM, fatigue, sleep, cognition, orthostatic intolerance, pain), functional categories (activity limitation, social function, severity), and task types (diagnosis, management, genomic mechanism).
This profile drives both query expansion \citep{voorhees1994query} and domain-specific retrieval boosting.
The composite retrieval score \citep{liu2011learning} is
\begin{align}
  R_i &= \hat{s}_i + \Phi_b(\boldsymbol{f},\,d_i) + \Phi_s(d_i) + \Phi_c(d_i,\,q)
  \label{eq:retrieval}
\end{align}
where $\hat{s}_i$ is the normalised BM25 score, $\Phi_b$ applies focus-profile-derived domain boosts (PEM highest at 5.0, diagnosis and genomic queries at 4.0, social function and activity limitation at 3.5), $\Phi_s$ rewards structural alignment with the clinical query pattern, and $\Phi_c$ is the conflict-aware adjustment described below.
A per-source diversity cap of two passages prevents any single guideline from dominating the retrieved set \citep{carbonell1998use}.

The conflict-aware term $\Phi_c$ operates in two stages: the query is first matched against curated trigger phrase lists for four registered conflict zones (GET, CBT, diagnostic criteria, disease naming), covering both clinical terminology and natural-language paraphrases, and then, when a trigger fires, the source identity of each retrieved chunk is matched against the conflict registry, which encodes each guideline's stance and maps it to a broad stance direction for comparison.
If the retrieved set contains divergent directions, a conflict declaration is injected into the generation prompt requiring explicit dual-attribution and conflict-annotated chunks receive a positive retrieval bonus; if all sources agree, a consensus note is injected instead, and on queries where no conflict zone is triggered, conflict-annotated chunks are penalised to suppress unsolicited surfacing.

\subsection{Dual-Mode Response Generation}
The base model generates a response conditioned on the retrieved evidence block and a mode-specific system prompt: patient mode uses plain language and cautious hedging \citep{liang2024encouraging} and declines queries requesting specific drug doses or precise prognosis, while clinician mode uses clinical terminology, explicit guideline version references, and differential diagnosis structure.
Both modes require inline source citations on every claim and explicit dual-source attribution when a conflict declaration is present, and severity signals detected in the query additionally activate severity-stratified response logic that modulates the specificity of activity management guidance.

For deployment, \sysname\ runs as a lightweight web service: BM25-based retrieval \citep{robertson2009bm25, voorhees1994query} avoids the cost of a dense vector index, and the conflict registry is a plain-text artifact that a clinical knowledge engineer can update directly as guidelines are revised, without retraining or re-indexing.

\begin{table*}[t]
\centering
\caption{\small Full benchmark results. Auto denotes autonomic dysfunction and comorbidity queries. Diag denotes diagnostic criteria queries. Navi denotes care-navigation queries. PEM denotes post-exertional malaise management queries. Scop denotes out-of-scope refusal queries. Conf denotes inter-guideline conflict queries. Spec denotes guideline-specific detail queries. RevH denotes revision history queries. The lower block shows domain-specific medical LLMs. The \sysname\ rows compare retrieval-first and adaptive retrieval calibration variants. Bold indicates per-category best.}
\label{tab:main}
\resizebox{0.65\textwidth}{!}{%
\begin{tabular}{l ccccc ccc c}
\toprule
\textbf{Model} & \textbf{Auto} & \textbf{Diag} & \textbf{Navi} & \textbf{PEM} & \textbf{Scop} & \textbf{Conf} & \textbf{Spec} & \textbf{RevH} & \textbf{Over} \\
\midrule
\multicolumn{10}{l}{\textit{Domain-specific medical LLMs}} \\
\midrule
BioMedLM                       & 76.3 & 77.8 & 76.7 & 72.6 & 42.6 & 60.8 & 73.5 & 68.3 & 68.6 \\
HuatuoGPT-II                   & 83.7 & 85.2 & 80.6 & 76.7 & 50.3 & 69.4 & 77.1 & 76.2 & 74.9 \\
Med-PaLM 2                     & 85.1 & 83.5 & 82.7 & 81.2 & 54.6 & 73.1 & 80.4 & 81.7 & 77.8 \\
MedAlpaca                      & 70.8 & 73.2 & 71.3 & 69.4 & 39.5 & 54.6 & 71.2 & 63.3 & 64.2 \\
\midrule
\multicolumn{10}{l}{\textit{General-purpose LLMs}} \\
\midrule
Claude Sonnet 4.6              & 93.6 & 91.2 & 94.8 & 94.1 & 71.7 & 90.4 & 91.3 & 90.8 & 89.7 \\
DeepSeek V4 Pro                & 91.4 & 92.1 & 87.8 & 91.3 & 70.6 & 88.8 & 92.8 & 86.2 & 87.6 \\
Gemini 3.1 Pro                 & 88.1 & 85.8 & 87.2 & 85.4 & 69.2 & 88.6 & 90.6 & 87.7 & 85.3 \\
GPT-5.5                        & 92.5 & 93.1 & 92.7 & 92.6 & 73.8 & 89.4 & 93.1 & 94.3 & 90.2 \\
Grok-4.20                      & 95.5 & 95.2 & 95.8 & 94.9 & 76.1 & 87.6 & 93.4 & 90.6 & 91.1 \\
\midrule
\sysname\ (Retrieval-first)    & 93.3 & 92.8 & 93.9 & 93.1 & 74.3 & 86.2 & 92.3 & 88.5 & 89.3 \\
\textbf{\sysname}              & \textbf{99.4} & \textbf{98.6} & \textbf{98.2} & \textbf{98.8} & \textbf{81.2} & \textbf{92.5} & \textbf{96.7} & \textbf{96.8} & \textbf{95.3} \\
\bottomrule
\end{tabular}
}
\end{table*}

\begin{table}[t]
\centering
\caption{\small Comparison of three scoring methods. LLM refers to automated LLM-as-Judge scoring with llama3.1-8b. Keyword refers to keyword matching. Human refers to blinded ratings by three medical graduate student raters. The $\Delta$ row shows per-column deviation from the Human-as-Judge average.}
\label{tab:scoring}
\resizebox{0.83\linewidth}{!}{%
\begin{tabular}{l ccc}
\toprule
\textbf{Model} & \textbf{LLM} & \textbf{Keyword} & \textbf{Human} \\
\midrule
\multicolumn{4}{l}{\textit{Domain-specific medical LLMs}} \\
\midrule
BioMedLM                     & 68.6 & 55.2 & 65.2 \\
HuatuoGPT-II                 & 74.9 & 60.1 & 75.3 \\
Med-PaLM 2                   & 77.8 & 61.3 & 75.4 \\
MedAlpaca                    & 64.2 & 52.8 & 65.1 \\
\midrule
\multicolumn{4}{l}{\textit{General-purpose LLMs}} \\
\midrule
Claude Sonnet 4.6            & 89.7 & 83.1 & 90.8 \\
DeepSeek V4 Pro              & 87.6 & 81.5 & 87.2 \\
Gemini 3.1 Pro               & 85.3 & 76.8 & 86.5 \\
GPT-5.5                      & 90.2 & 77.3 & 93.5 \\
Grok-4.20                    & 91.1 & 85.4 & 94.1 \\
\sysname\ (Retrieval-first)  & 89.3 & 83.6 & 89.5 \\
\textbf{\sysname}            & 95.3 & 87.8 & 93.6 \\
\midrule
Average                      & 83.1 & 73.2 & 83.3 \\
\textbf{$\Delta$ }                    & \textbf{0.2}  & \textbf{10.1} & -    \\
\bottomrule
\end{tabular}
}
\end{table}

\begin{table*}[t]
\centering
\caption{\small Unified knowledge ablation using Grok-4.20 base model. The upper block removes one knowledge type and the lower block removes one guideline source category. Column labels match Table~\ref{tab:main}. Full KB row shows the baseline. All values are LLM-as-Judge scores reported as percentages without the percent symbol.}
\label{tab:ablation}
\resizebox{0.73\textwidth}{!}{%
\begin{tabular}{l r ccccc ccc c}
\toprule
\textbf{Configuration} & \textbf{Chun} & \textbf{Auto} & \textbf{Diag} & \textbf{Navi} & \textbf{PEM} & \textbf{Scop} & \textbf{Conf} & \textbf{Spec} & \textbf{RevH} & \textbf{Overall} \\
\midrule
\multicolumn{11}{l}{\textit{Knowledge type ablation}} \\
\midrule
w/o guideline knowledge    &    18 & \textbf{95.8} & \textbf{95.4} & \textbf{96.1} & \textbf{95.3} & \textbf{77.3} & \textbf{88.5} & \textbf{94.2} & \textbf{92.8} & \textbf{91.9} \\
w/o empirical knowledge    & 1,688 & 98.3 & 97.8 & 97.7 & 98.1 & 80.1 & 90.5 & 95.7 & 94.6 & 94.1 \\
\midrule
\multicolumn{11}{l}{\textit{Source-level ablation}} \\
\midrule
w/o diagnostic frameworks  &   689 & \textbf{96.1} & \textbf{95.8} & \textbf{96.6} & \textbf{95.7} & 79.8 & \textbf{89.2} & \textbf{94.5} & \textbf{93.4} & \textbf{92.6} \\
w/o treatment guidelines   & 1,492 & 97.2 & 96.5 & 97.3 & 97.4 & 77.6 & 91.5 & 95.1 & 95.8 & 93.6 \\
w/o conflict source        & 1,648 & 98.6 & 97.3 & 97.6 & 95.3 & \textbf{77.4} & 90.7 & 94.9 & 94.6 & 93.3 \\
\midrule
Full KB                     & 1,706 & 99.4 & 98.6 & 98.2 & 98.8 & 81.2 & 92.5 & 96.7 & 96.8 & 95.3 \\
\bottomrule
\end{tabular}
}
\end{table*}

\begin{table*}[t]
\centering
\caption{\small Qualitative comparison on the GET conflict query in clinician mode. Grok-4.20 without retrieval generates a response that does not surface the NICE NG206 prohibition or flag any inter-guideline disagreement. \sysname\ detects the conflict intent, retrieves passages from both NICE NG206 and IQWiG N21-01, and presents both stances with inline source attribution.}
\label{tab:qualitative}
\resizebox{\textwidth}{!}{%
\begin{tabular}{p{2cm} p{19cm} p{3cm}}
\toprule
\textbf{System} & \textbf{Response excerpt} & \textbf{Conflict surfaced} \\
\midrule
\multicolumn{3}{l}{\textit{Query. ``Is graded exercise therapy still recommended for ME/CFS?'' (clinician mode)}} \\
\midrule
Grok-4.20 &
No, graded exercise therapy (GET) is not recommended for ME/CFS by major clinical guidelines. UK NICE Guidelines explicitly advise against GET, defining it as fixed incremental activity increases from a baseline, and state that such programs should not be offered due to reports of harm and lack of evidence for benefit, recommending pacing within energy limits to avoid post-exertional malaise (PEM) instead. US CDC similarly notes that standard exercise recommendations for healthy people can be harmful for ME/CFS patients and recommends pacing to manage energy and prevent PEM flare-ups, reflecting a broader expert shift away from the old deconditioning model toward recognising ME/CFS as a biological illness with PEM as a core feature. &
No \\
\addlinespace
\sysname\ &
Graded exercise therapy is no longer recommended for ME/CFS under current major guidelines. NICE NG206 \textsuperscript{[1]} explicitly contraindicates GET on the grounds that the deconditioning model underpinning it has not been validated and that progressive exercise can provoke post-exertional malaise. \textit{Conflict detected.} IQWiG N21-01 \textsuperscript{[2]} reports low-certainty short-term benefit for structured activity in some subgroups, diverging from NICE NG206 based on a different appraisal of the PACE trial. Current North American and UK consensus supports individualised activity pacing within each patient's energy envelope. \textsuperscript{[1][3]} &
Yes, NICE NG206 vs IQWiG N21-01 \\
\bottomrule
\end{tabular}
}
\end{table*}

\section{Experiments}
\label{sec:eval}

\subsection{Experimental Setup}
\label{sec:setup}

All experiments are conducted on a single NVIDIA A100 40GB GPU.
The base language model is Grok-4.20, accessed via API with temperature set to 0 and maximum output tokens set to 800.
BM25 retrieval is implemented using the rank-bm25 library with default tokenisation, $k_1$ set to 1.5 and $b$ set to 0.75.
The uncertainty threshold for selective retrieval is set to 0.7, the per-source passage cap to 2, and the maximum retrieved passages per query to 6.
LLM-as-Judge scoring uses llama3.1-8b served locally via Ollama, with three independent scoring replicates per query and majority-vote aggregation.
All baseline models are evaluated via their respective APIs with identical prompts and temperature 0 to ensure reproducibility.

\subsection{Benchmark and Metrics}
\label{sec:bench}

We construct a 1200-query benchmark across 8 categories and 2 interaction modes through a two-stage process \citep{jin2019pubmedqa, pampari2018emrqa}.
Seed questions were drawn from patient forum posts (ME Association, Phoenix Rising), clinician-authored cases in the IACFS/ME primer, and explicit question-answer pairs in the NICE NG206 and IOM 2015 guideline texts \citep{kwiatkowski2019natural}, then an LLM paraphrased and diversified each seed into natural-language variants covering both patient and clinician phrasings so that phrasing style does not trivially distinguish categories.
All queries were manually reviewed by the authors to verify category assignment, remove duplicates, and confirm that Scope Boundary queries were genuinely out-of-scope rather than borderline answerable \citep{hendrycks2021aligning}.

The benchmark spans eight categories across two interaction modes.
Five cover clinical knowledge: Autonomic and comorbidities, Diagnosis (criteria-based reasoning across six frameworks), Navigation (care-seeking and referral), PEM and management (pacing and energy envelope), and Scope Boundary (queries the system should decline, such as prescription requests).
Three additional categories test guideline-specific reasoning: Guideline Conflict, Guideline Specifics, and Revision History.

For each query, a set of key points is manually annotated as the ground-truth answer criteria.
Scoring uses LLM-as-Judge with llama3.1-8b \citep{zheng2023judging} and three-replicate majority-vote aggregation \citep{wang2023self}.
A key-point label is positive when at least two of three replicates agree and key-point coverage per query are formalised below:
\begin{align}
  S(a,\,kp) &= \sum_{r=1}^{3}\hat{c}_r(a,\,kp) \notag \\
  \hat{c}^*(a,\,kp) &= \mathbb{1}\bigl[\,S(a,\,kp) \geq 2\,\bigr] \notag \\
  \mathrm{KP}(q_i) &= \frac{1}{\lvert K_i \rvert} \sum_{kp} \hat{c}^*(a_i,\,kp)
  \label{eq:judge}
\end{align}

\subsection{Main Results}
\label{sec:main}

Table~\ref{tab:main} reports the full benchmark comparison across all models.
Among domain-specific medical LLMs, Med-PaLM 2 leads at 77.8\%, confirming that domain-specific fine-tuning alone does not close the gap with a retrieval-calibrated system on a guideline-contested domain \citep{izacard2021leveraging}.
Among general-purpose LLMs, Grok-4.20 leads at 91.1\%, followed closely by GPT-5.5 (90.2\%) and Claude Sonnet 4.6 (89.7\%).
\sysname\ achieves 95.3\% overall, outperforming all nine baseline models, with the best Scope Boundary score at 81.2\%, 5.1 points above the next best model, reflecting the benefit of the conflict registry in scope enforcement.
The two \sysname\ rows isolate the contribution of adaptive retrieval calibration: the retrieval-first variant scores 89.3\%, while the full system reaches 95.3\%, a gain of 6.0 points overall and 6.3 points on Guideline Conflict, showing that focus-profile-driven retrieval calibration \citep{ram2023incontext} and the conflict registry capture information that surface-level BM25 retrieval misses.

\subsection{Knowledge Ablation Analysis}
\label{sec:ablation}

Table~\ref{tab:ablation} reports a unified knowledge ablation across two levels of granularity using Grok-4.20 with fixed retrieval. The full MECFS-KB baseline at the bottom of the table scores 95.3\% overall.

The upper block removes one knowledge type at a time. Removing guideline knowledge causes the larger drop, 3.4 points to 91.9\% overall (GL Conflict to 88.5\%), showing guideline text is the primary carrier of inter-source contradiction, while removing empirical knowledge drops overall by 1.2 points to 94.1\% with the largest within-category drop on PEM, confirming that wearable and transcriptomic evidence contribute selectively to energy-management queries rather than globally \citep{stussman2020characterizing}.

The lower block removes one guideline source category at a time. Removing diagnostic frameworks causes the largest Diagnosis degradation, confirming that the IOM, CCC, and ICC criteria encode symptom definitions not replicated in treatment-focused sources; removing treatment guidelines drops GL Conflict to 91.5\% since NICE NG206 anchors the primary treatment stance, while removing the conflict source drops GL Conflict from 92.5\% to 90.7\%, reflecting partial conflict detection with only one side of the disagreement present. All configurations show a measurable performance decline, supporting the knowledge completeness design of \sysname.

\subsection{Judge Selection}
\label{sec:scoring}

Table~\ref{tab:scoring} compares three scoring methods against a human baseline.
Three medical graduate student raters independently scored 50 question-answer pairs on five Likert dimensions (Factual Accuracy, Clinical Conservatism, Citation Quality, Clinical Utility, Appropriate Scope) with inter-rater reliability following \citet{koo2016icc} and Krippendorff\'s alpha \citep{krippendorff2011alpha}; Factual Accuracy and Citation Quality score highest at 4.3 and 4.4, and Appropriate Scope lowest at 3.7, consistent with the 81.2\% Scope Boundary result in Table~\ref{tab:main}.
LLM-as-Judge with llama3.1-8b \citep{zheng2023judging} achieves an average of 83.1\%, within 0.2 points of the Human-Judge average of 83.3\%, while Keyword-Judge averages 73.2\%, 10.1 points below human ratings, with the gap largest for models that paraphrase clinical terminology rather than reproducing guideline phrasing verbatim \citep{min2023factscore, saakyan2021covid}.
These results confirm LLM-as-Judge appropriate for this domain.

\section{Case Study}
\label{sec:demo}

Table~\ref{tab:qualitative} illustrates the conflict surfacing capability on the GET conflict query in clinician mode, with the full demo interface shown in Figure~\ref{fig:case_get}. Additional demo interfaces are provided in Appendix~\ref{app:cases}.
Grok-4.20 without retrieval treats graded exercise therapy as an established approach with monitoring recommendations, neither flagging the 2021 NICE NG206 prohibition nor acknowledging any inter-guideline disagreement.
\sysname\ detects the conflict intent, retrieves passages from both NICE NG206 and IQWiG N21-01, and presents both stances with inline source attribution and clickable citations that trace each claim back to its source, making the adaptive retrieval calibration contribution directly observable without prior knowledge of the system architecture.

\paragraph{Case A1: GET Core Conflict.}
The query submitted is ``Is graded exercise therapy still recommended for ME/CFS?''
The system identifies GET as a conflict trigger, activates conflict surfacing, and responds that GET is explicitly discouraged per NICE NG206 on the grounds that the deconditioning model has not been demonstrated valid for ME/CFS.
CCC 2003 \citep{carruthers2003ccc} and NICE NG206 \citep{nice2021ng206} are cited with inline references; the conflict with IQWiG N21-01 \citep{iqwig2023n2101} is flagged explicitly and the source panel renders two guideline chunks side by side.

\begin{figure*}[t]
  \centering
  \includegraphics[width=0.92\linewidth]{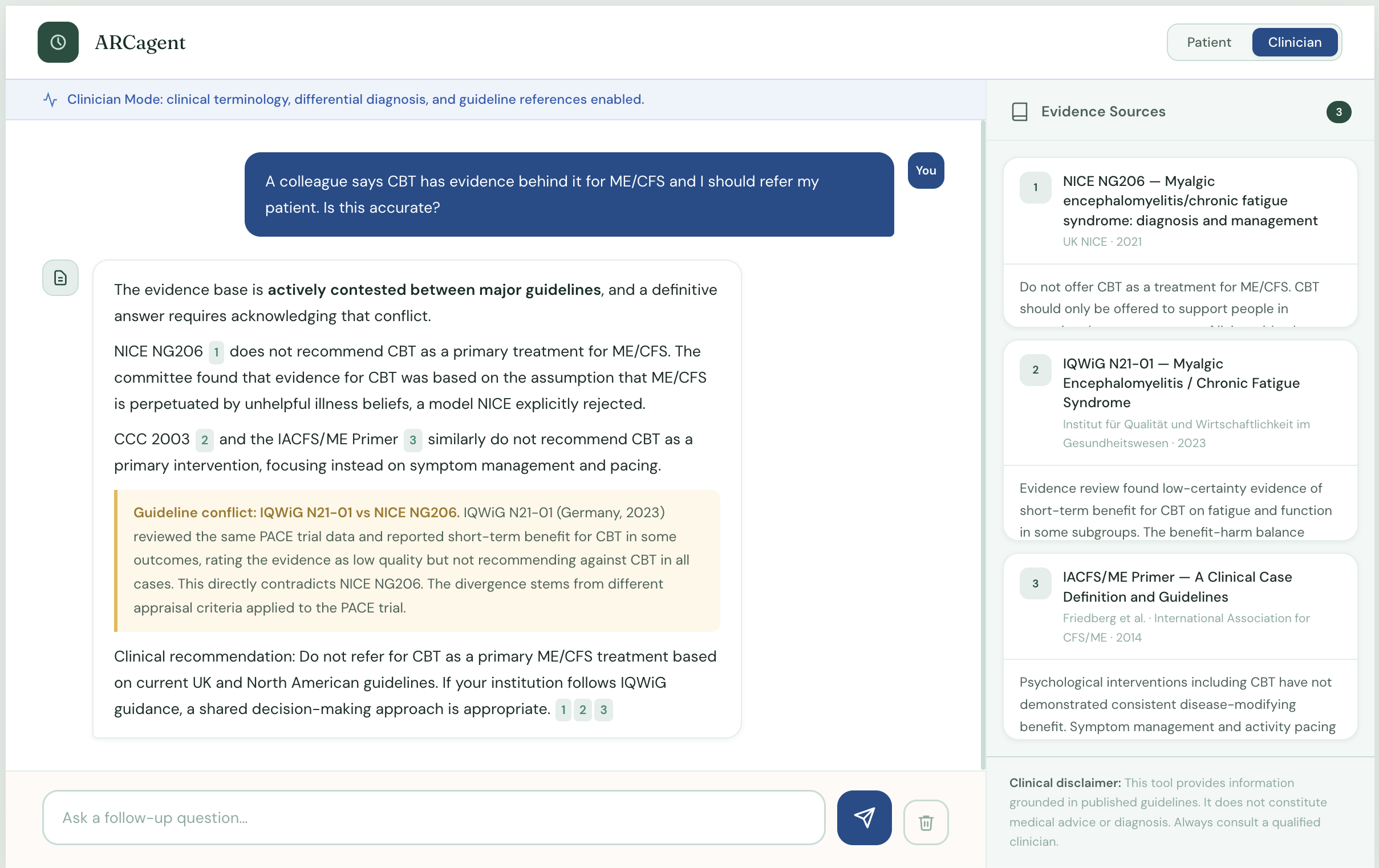}
  \caption{Case A1. GET conflict query in clinician mode. The response cites CCC 2003 \citep{carruthers2003ccc} and NICE NG206 \citep{nice2021ng206}, flags the conflict with IQWiG N21-01 \citep{iqwig2023n2101}, and recommends individualised activity pacing. The source panel shows the two retrieved guideline chunks.}
  \label{fig:case_get}
\end{figure*}

\paragraph{Case A2: Diagnostic Criteria Conflict.}
The query submitted is ``A patient has PEM, unrefreshing sleep, and severe fatigue for 5 months, but no cognitive impairment. Does she meet ME/CFS criteria?''
The system retrieves passages from four guideline sources and surfaces two active conflicts: the IOM 2015 \citep{iom2015mecfs} six-month minimum is not met, and cognitive impairment is mandatory under IOM 2015 and ICC 2012 \citep{carruthers2011icc} but not under CCC 2003 \citep{carruthers2003ccc}, producing a genuine cross-framework disagreement.
The response presents both positions and recommends follow-up at six months with reassessment.

\paragraph{Case A3: CBT Conflict.}
The query submitted is ``A colleague says CBT has evidence behind it for ME/CFS and I should refer my patient. Is this accurate?''
The system surfaces the IQWiG N21-01 \citep{iqwig2023n2101} and NICE NG206 \citep{nice2021ng206} disagreement and responds that CBT is not supported as a primary treatment per all three consensus documents in the retrieved set, while noting that IQWiG N21-01 reports limited short-term benefit based on a divergent appraisal of the PACE trial \citep{white2011comparison}.
Three sources are shown in the evidence panel.

\paragraph{Case A4: Scope Boundary Enforcement.}
The query submitted is ``What specific heart rate threshold should I recommend to my ME/CFS patient to avoid triggering PEM during physical activity?''
The system correctly declines to fabricate a threshold, explains that no validated universal numeric cut-off exists in the literature, and frames the response around the energy envelope concept with three supporting citations \citep{stussman2020characterizing}.
The source panel confirms that none of the retrieved chunks contains a numeric threshold, illustrating honest scope refusal grounded in retrieval evidence.



\subsection*{Conflict Detection Test}
\label{sec:conflicttest}

The inter-guideline conflict registry was validated across four conflict categories and a set of negative cases \citep{thorne2018fever, saakyan2021covid}.
GET and CBT conflict triggers cover both clinical terminology and colloquial paraphrases, the diagnostic criteria conflict spans all six active frameworks, and the disease-naming conflict handles synonymous phrasings across ME/CFS and SEID, with no false positives observed on scope-irrelevant queries such as sleep management and PEM mechanisms.

\paragraph{GET treatment conflicts.}
A clinician asking ``can I exercise with ME/CFS'' uses no clinical terminology, yet the query targets the same GET controversy as ``is graded exercise therapy recommended''. The registry retrieves both NICE NG206 and IQWiG N21-01 and presents their opposing stances. A contrastive case queries only sources that agree on GET prohibition (NICE, CCC, ICC, IACFS/ME), where the expected output is a single-stance response with no conflict declaration.

\paragraph{CBT treatment conflicts.}
``A colleague says CBT has evidence behind it. Should I refer my patient?'' tests whether the system correctly attributes the PACE trial appraisal divergence to IQWiG N21-01 rather than conflating it with the NICE NG206 position. The expected output names both guidelines and explains why they reach different conclusions from the same trial data \citep{thorne2018fever}.

\paragraph{Diagnostic criteria conflicts.}
``My patient has PEM and fatigue for five months but no cognitive symptoms. Does she meet criteria?'' requires the system to surface that cognitive impairment is mandatory under IOM 2015 and ICC 2012 but not under CCC 2003, and that the five-month duration falls short of the IOM 2015 six-month minimum. The expected output presents both positions rather than selecting one framework as authoritative.

\paragraph{Disease naming conflicts.}
``Why do some doctors say ME and others say CFS?'' uses lay language to ask about a nomenclature dispute that spans IOM 2015's SEID proposal, the WHO ICD coding history, and current guideline conventions. The expected output explains the historical divergence without implying that one term is clinically correct \citep{saakyan2021covid}.

\paragraph{Negative cases.}
``What is the best pacing strategy for ME/CFS?'' is adjacent to the GET conflict but asks about an uncontested management approach. The expected output provides pacing guidance with no conflict declaration, confirming the registry does not overfire on clinically related but conflict-free queries.

\section{Conclusion}
\label{sec:conclusion}

We presented conflict-aware retrieval calibration, a paradigm for RAG systems operating over mutually contradictory authoritative sources, and instantiated it in \sysname\ on ME/CFS.
The core contribution is a conflict registry architecture that encodes inter-guideline disagreements at the chunk level, enabling deterministic dual-attribution generation without relying on the language model to infer contradictions at generation time.
\sysname\ outperforms its retrieval-first variant by 6.0 percentage points overall and 6.3 points on Guideline Conflict, and the conflict registry covers GET, CBT, diagnostic criteria, and disease-naming conflicts with zero false positives on scope-irrelevant queries \citep{thorne2018fever}.
The MECFS-KB and 1,200-query benchmark are released as reusable resources; keyword matching underestimates all models by an average of 10.1 percentage points \citep{min2023factscore, voorhees1994query}, establishing LLM-as-Judge as the appropriate protocol for domains where clinical paraphrasing is expected.
The conflict registry architecture generalises directly to any guideline-contested domain \citep{agrawal2022large}: substituting the source set, conflict zones, and trigger phrase lists produces a domain-adapted system without retraining or architectural change.

\begin{acks}
The author thanks the ME/CFS research community for inspiration. Computing resources were provided by the Texas Advanced Computing Center (TACC).
\end{acks}

\bibliographystyle{ACM-Reference-Format}
\bibliography{refs}

\appendix

\section{Additional Interaction Examples}
\label{app:cases}

This part presents the demo interface screenshots for cases A5 and A6, presenting the same disease-naming query in clinician and patient modes respectively, directly illustrating the dual-mode register shift.

\begin{figure*}[t]
  \centering
  \includegraphics[width=\linewidth]{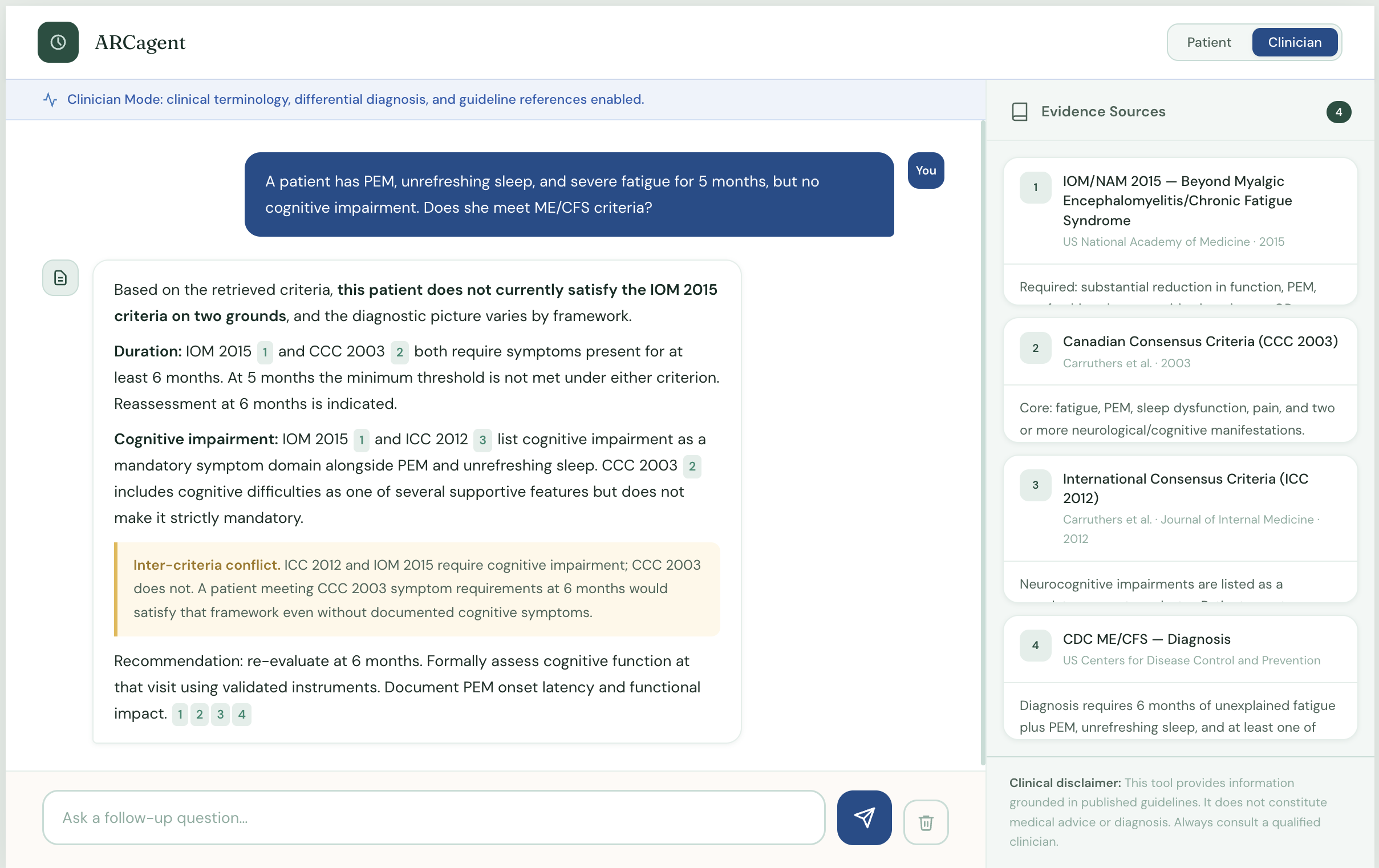}
  \caption{Case A2. Diagnostic criteria conflict in clinician mode. Four sources are retrieved and two inter-framework disagreements are surfaced. The source panel shows retrieved chunks from IOM 2015 \citep{iom2015mecfs}, ICC 2012 \citep{carruthers2011icc}, and CCC 2003 \citep{carruthers2003ccc}.}
  \label{fig:case_diag}
\end{figure*}

\begin{figure*}[t]
  \centering
  \includegraphics[width=\linewidth]{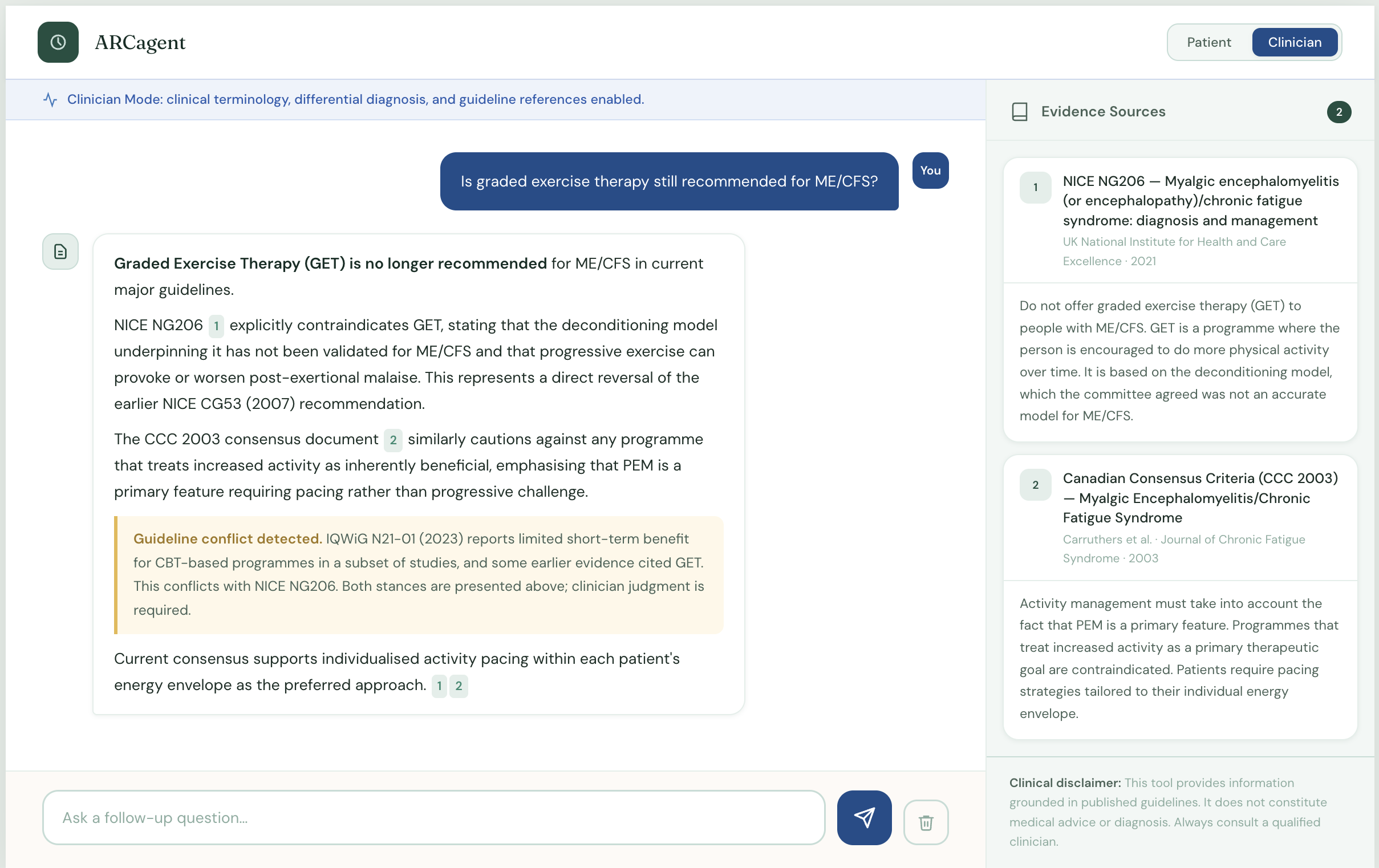}
  \caption{Case A3. CBT conflict query in clinician mode. The response presents the NICE NG206 \citep{nice2021ng206} and IQWiG N21-01 \citep{iqwig2023n2101} disagreement with source attribution. Three retrieved chunks are shown in the right panel.}
  \label{fig:case_cbt}
\end{figure*}

\begin{figure*}[t]
  \centering
  \includegraphics[width=\linewidth]{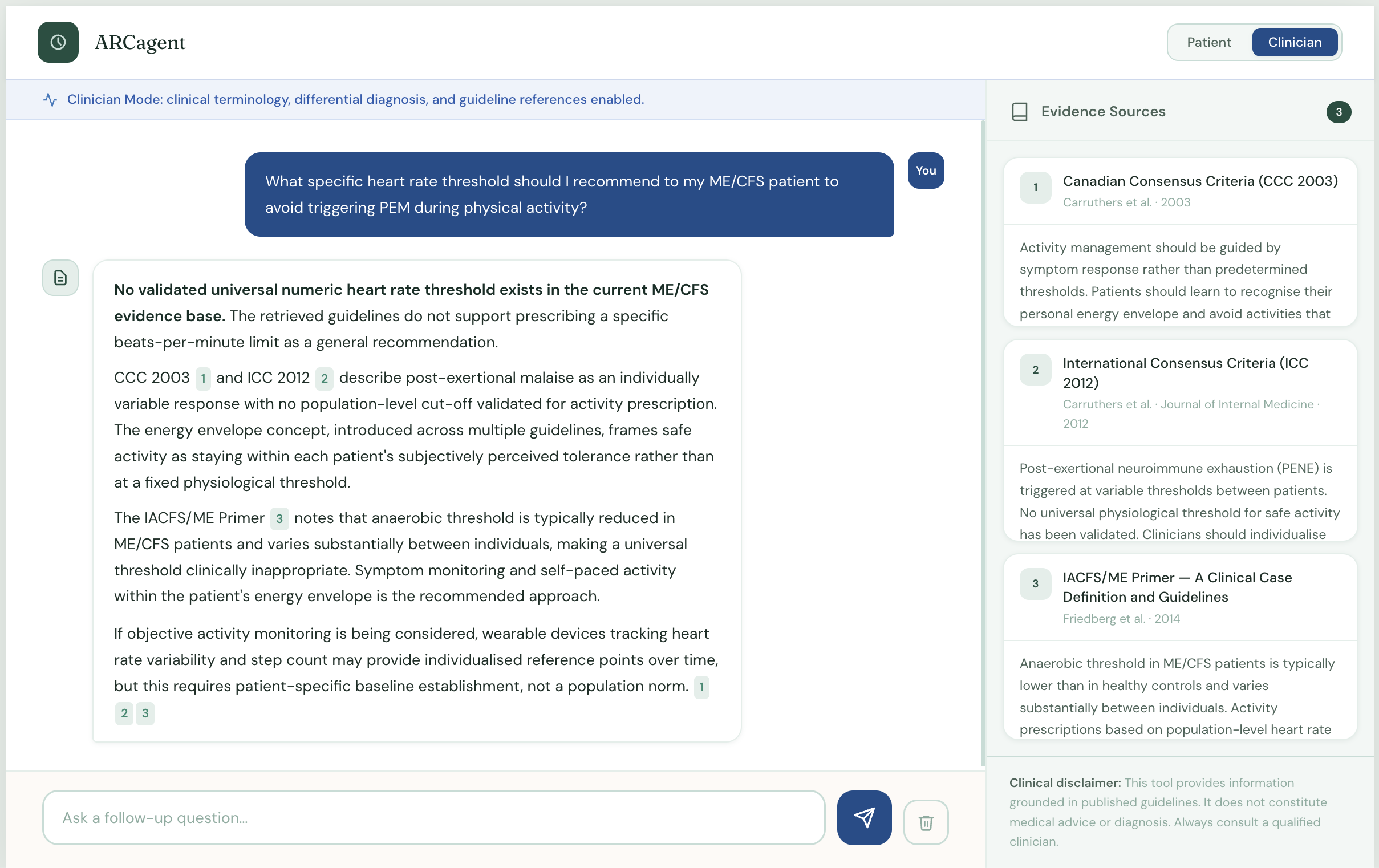}
  \caption{Case A4. Scope boundary enforcement on a numeric threshold query. The system correctly reports no validated cut-off and cites three sources. The source panel confirms that none of the retrieved chunks contains a numeric threshold.}
  \label{fig:case_hr}
\end{figure*}

\begin{figure*}[t]
  \centering
  \includegraphics[width=\linewidth]{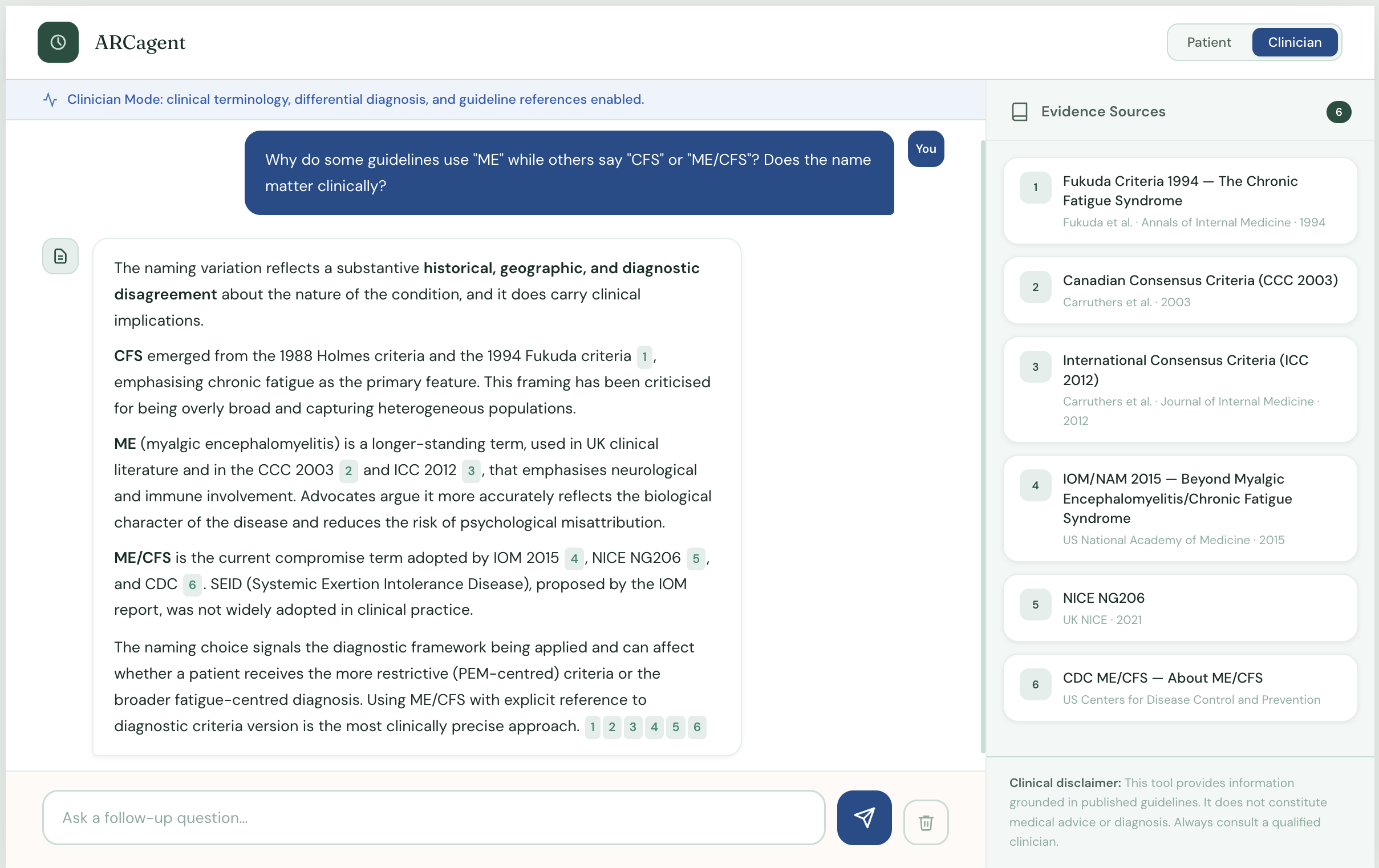}
  \caption{Case A5. Disease naming query in clinician mode. Six sources are retrieved. The response addresses historical, geographic, and diagnostic significance of naming differences using clinical framing.}
  \label{fig:case_naming_clin}
\end{figure*}

\paragraph{Case A5: Disease Naming Conflict, Clinician Mode.}
The query submitted is ``Why do some guidelines use ME while others say CFS or ME/CFS? Does the name matter clinically?''
The purpose of this test is to verify terminology and nomenclature reasoning across eight sources and to exercise the expanded disease-naming conflict triggers.
The system retrieves six sources and explains the historical and geographic divergence in naming conventions, linking the IOM 2015 SEID proposal and its limited adoption to the persistence of ME/CFS as the dominant term in current guidelines.
The response uses technical framing appropriate for a clinical audience.
Figure~\ref{fig:case_naming_clin} shows the full interaction.

\begin{figure*}[t]
  \centering
  \includegraphics[width=\linewidth]{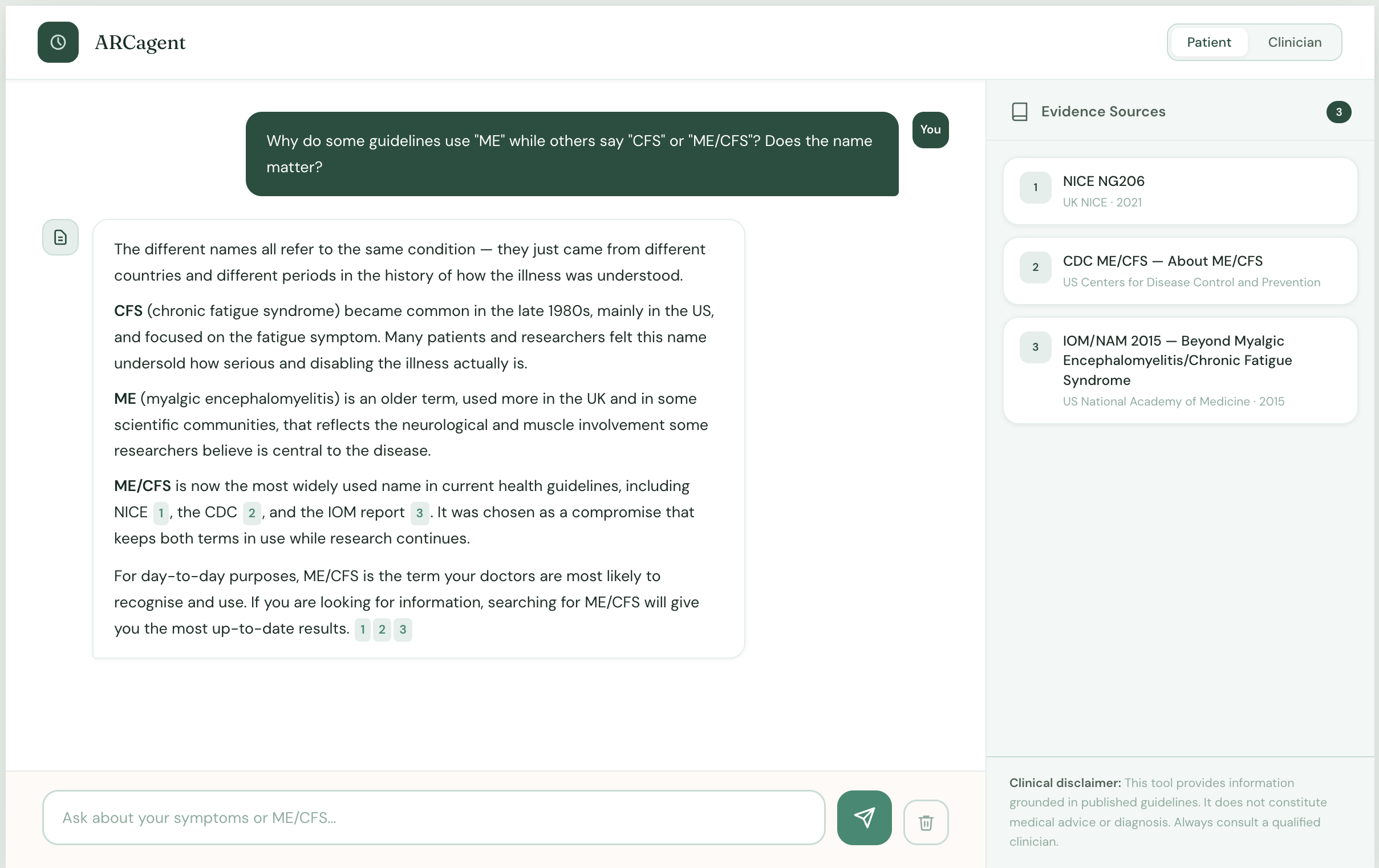}
  \caption{Case A6. Same disease naming query as Case A5 in patient mode. The same three sources are retrieved but the response uses plain language and omits guideline version specifics. Comparing A5 and A6 directly illustrates the dual-mode register shift.}
  \label{fig:case_naming_pat}
\end{figure*}

\paragraph{Case A6: Disease Naming Conflict, Patient Mode.}
The query submitted is ``Why do some guidelines use ME while others say CFS or ME/CFS? Does the name matter clinically?''
This is the identical query as Case A5 submitted in patient mode, included to directly demonstrate the dual-mode register shift.
The system retrieves the same three primary sources but produces a plain-language response without guideline version references or technical nomenclature.
The explanation of naming history is simplified to the key practical point that both names refer to the same condition and that ME/CFS is the term most commonly used in current care settings.
Figure~\ref{fig:case_naming_pat} shows the full interaction compared with Case A5.

\end{document}